# Classification of Diabetic Retinopathy Images Using Multi-Class Multiple-Instance Learning Based on Color Correlogram Features

Ragav Venkatesan, Parag Chandakkar, Baoxin Li, *Senior Member, IEEE*, and Helen K. Li, MD

*Abstract*— All people with diabetes have the risk of developing diabetic retinopathy (DR), a vision-threatening complication. Early detection and timely treatment can reduce the occurrence of blindness due to DR. Computer-aided diagnosis has the potential benefit of improving the accuracy and speed in DR detection. This study is concerned with automatic classification of images with microaneurysm (MA) and neovascularization (NV), two important DR clinical findings. Together with normal images, this presents a 3-class classification problem. We propose a modified color auto-correlogram feature (AutoCC) with low dimensionality that is spectrally tuned towards DR images. Recognizing the fact that the images with or without MA or NV are generally different only in small, localized regions, we propose to employ a multi-class, multiple-instance learning framework for performing the classification task using the proposed feature. Extensive experiments including comparison with a few state-of-art image classification approaches have been performed and the results suggest that the proposed approach is promising as it outperforms other methods by a large margin.

## I. INTRODUCTION

Diabetic retinopathy (DR) is a common cause of blindness among the diabetic population. Despite various advances in diabetes care over the years, loss of vision is still a potentially devastating complication in people with diabetes. The risk of severe vision loss can be reduced significantly by timely diagnosis and treatment of DR. Maximizing the efficiency and accuracy of assessing DR severity levels could help prevent vision disabilities and their resulting high cost to the society.

The conventional process of evaluating retinal fundus images in DR diagnosis is laborious and prone to error or reviewer fatigue. Recent years have seen many research efforts on developing computer-assisted detection and evaluation of diabetic retinal lesions [1, 2, 3, 4, 5]. While progresses have been made, the lack of a unified and systematic solution or system that has been widely accepted in ophthalmology indicates that the central problems have not been solved. For example, a content-based retireval system for retinal images is discussed in [3], where the retrieved images appear to be largely similar in appearance but not similar in terms of clinical relevance, and thus the practical usefulness of the approach is unclear.

Two important DR clinical findings are microaneurysm (MA) and neovascularization (NV). Among others (e.g., intra-retinal hemorrhages, exudates, etc.), MA is characteristic of non-proliferate diabetic retinopathy (NPDR) or background retinopathy. On the other hand, the early proliferate diabetic retinopathy (PDR) stage is characterized by neovascularization (NV), which is the formation of abnormal new blood vessels. Therefore, MA and NV are two clinically important lesion types to consider. In this study, we focus on the following 3-category classification problem: classifying a given DR image as one of the three types, a normal image (no DR), an image with MA, or an image with NV. Given the aforementioned importance of MA and NV in DR diagnosis, such a 3-category classification problem has the potential of contributing to developing computer-based systems for delivering clinically relevant results.

Image classfication is a well-studied topic in the field of computer vision. State-of-the-art approaches include those relying on robust features and classfication algorithms that have been developed in the past decade. For example, the scale-invariant feature transform (SIFT) features coupled with a bag-of-words (BoW) approach [6] has been shown to be very effective. In such an approach, distinctive and repeatable image features like the differene-of-Gaussian (DoG) points are detected. An image patch for each feature is extracted from which a feature descriptor is then computed. The descriptors are further encoded into a visual word via some learned codebook. Such features can then be used by classifiers such as support vector machines (SVM) to perform image classification. Such approaches are in general inadeqaute for our 3-category classification problem due to two factors: (i) the BoW features are mostly global and thus lacking the desired discriminating local features which are critical for DR image classificaiton; and (ii) the difference among images of the three categories under consideration often boils down to only localized regions of the images, rendering an SVM classifer ineffective when it is applied to a global feaure. The second factor is further complicated by the fact that often the images are labeled on a per-image basis (i.e, without exact information regarding which regions define the label of the an image). This prevents the direct application of SVM on a region basis.

In this paper, we develop a spectrally-tuned color auto-correlogram and use it as the feature for classfication. Color correlograms [7] implicitly describe the global correlation of local spatial correlation of colors. This is a strong representation of textures with considerably small dimensionality. The proposed method further enhances such strength by spectrally tuning such features based on the observation that all DR images are relatively saturated in the red channel. Moreover, recognizing the fact that the images with or without MA or NV are in general only different in

R. Venkatesan, P. Chandakkar, and B. Li are with the Arizona State University, Tempe, AZ, 85281 USA (phone: 480-965-1735; fax: 480-965-2751; e-mail: baoxin.li@asu.edu).

H. K. Li is with Weill Cornel Medical College/The Methodist Hospital, The University of Texas Health Science Center Houston, and Thomas Jefferson University. (e-mail: hli@commuityretina.com).

small, localized regions, and that the images are in general only labeled at the per-image level, we propose to employ a multi-class, multiple-instance learning (MIL) [8] framework for performing the classification task using the proposed feature. Together, the development of a spectrally-tuned auto-correlagram and the employment of an MIL scheme lead to a novel classfication approach that is able to overcome typical challegens in DR image classification.

The remainder of this article is organized as follows: Section 2 describes the proposed algorithm. Section 3 provides the experimental results, including comparison with other approaches. Section 4 provides concluding remarks.

## II. PROPOSED APPROACH

The color correlogram of an image as proposed in [7] is a table indexed by color pairs, where the $k^{th}$ entry for the color pair $(i,j)$ specifies the probability of finding a pixel of color $j$ at a distance $k$ from a pixel of color $i$. Let $I$ be a square image of side $N$. Typically, to reduce the dimensionality of the features, the image is quantized into $m$ color bins, say $c_1, c_2, \ldots c_m$. Let $C(p)$ denote the color of a pixel $p = (x, y) \in I$. An $L_\infty$-norm is then caluclated to measure the distance between two pixels, say $p_1 = (x_1, y_1)$, $p_2 = (x_2, y_2)$, with $|p_1 - p_2| = \max(|x_1 - x_2|, |y_1 - y_2|)$. The histogram of the image $I$ (denoted by $h_i$) with respect to each bin is now calculated. This is equivalent to calculating the number of pairs of pixels, such that $|p_1 - p_2| = k$ and dividing it by the total number of pixels belonging to that particular bin in the entire image. We limit our search neighborhood to $3X3$. Also, we analyse the distribution of only those pixels which lie in the same bin $(i = j)$. Thus the features are called color auto-correlogram (AutoCC). This also ensures that the number of bins in an image is also the dimensionality of the feature space as the histogram of correlations of the bins becomes the feature vector.

To restrict the dimensionality of AutoCC features to 44, Li in [9] proposed a quantization scheme modeled after human vision, taking into account the color spectrum of natural images as shown in Fig. 1. However, as illustrated in Fig. 2, where we can observe that the spectrum for DR images is significantly different from that of the natural images considered in [9]. It can be noticed from Fig. 2 that the DR images have a saturated spectrum in the red channel. For AutoCC features to work on their best, we require uniformly shaded bins. To this end, we introduce histrogram equalization in the red channel, before performing any feature extraction. Effectively, this results in a spectrally-tuned set of features. We elaborate this in the following subsections, with illustration on the benefits of the spectrally-selective tuning. A classifier is then trained based on the extracted features from a training set.

### A. Quantizer design

All the images in the training set are considered and the red channels of the images are equalized. Once equalization is performed, we convert the training images into a 64-bin non-uniformly quantized one-channel image. The choice of the 64-bins was empirically determined from a codebook generation scheme using *k*-means clustering, as described below.

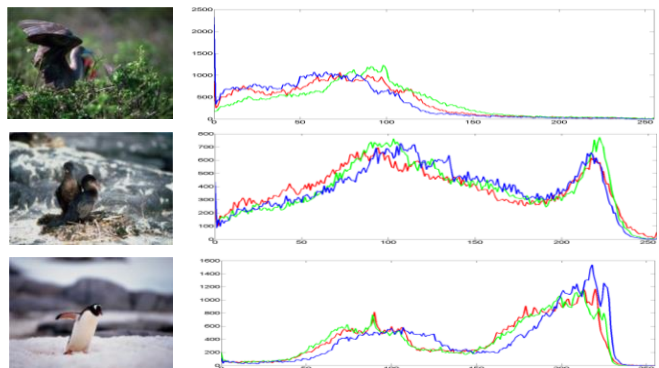

**Figure 1.** Histograms of some natural images. The variation in the histogram is clearly noticed within the three sample images.

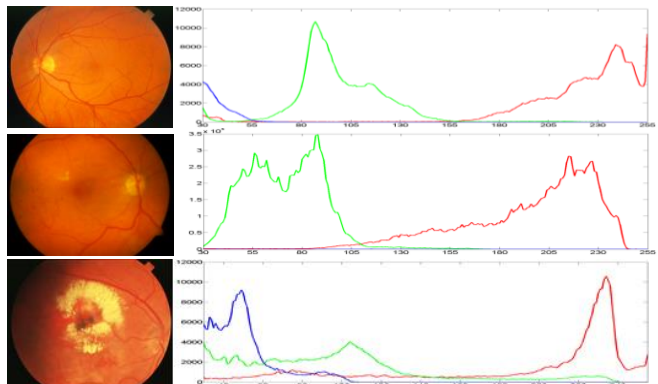

**Figure 2.** Histograms of typical fundus images. The top image is diagnosed as a normal image, the middle image as MA and the bottom image NV. It can be noticed clearly how the red channel for all the classes are saturated.

We first extract all the unique shades (<R,G,B> triplets) in the training set. Once all the unique shades present in the database are extracted, they are arranged in a $NX3$ matrix. Clustering based on *k*-means is then performed on the above mentioned matrix. The centroids produced from the *k*-means are the codebook of quantization bins. Once the codebook is generated, feature extraction can be performed.

Figures 3 and 4 show a visualization of this clustering approach for the quantization scheme proposed in [9] and the new approach respectively. It can be noticed that while the approach in Fig. 3, designed for natural images, has a varied density of points associated with each codeword, the proposed approach has a uniform density. Though the codewords are uniformly distributed in the color space, densities of shades are non-uniform in Fig. 3, while in the proposed approach, though the codewords are distributed non-uniformly, the associated shades per codeword are uniform. This results in reduced entropy of the feature space and also utilizes the bandwidth provided by the 64 dimensions to full extent in the proposed feature space. It can also be observed that most of the clusters in the original approach, though uniformly placed, are empty. This wastes many dimensions in the original AutoCC, while the proposed approach utilizes the 64 dimensions more efficiently. This also implies that the semantic information useful for a later classifier learning stage is distributed evenly along the 64 dimensions, which is a desirable property to have for a feature space.

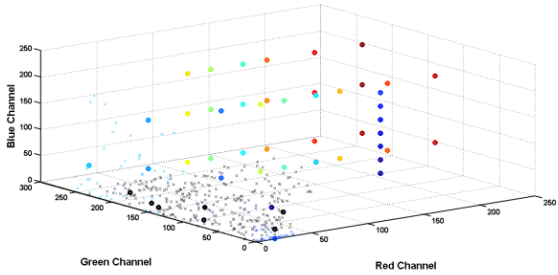

**Figure 3.** Quantization of AutoCC approach proposed in [9].

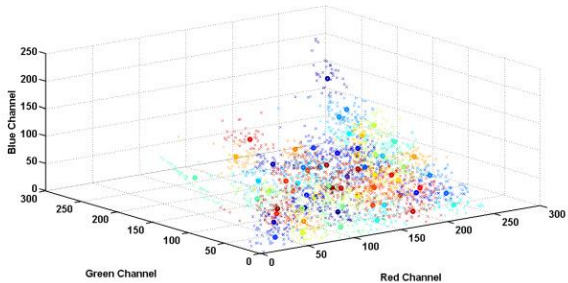

**Figure 4.** Quantization of the proposed AutoCC approach.

*B. Feature extraction and classifier design*

With the quantization scheme established above, features are extracted to support a multiple instance learning (MIL) approach. As was already discussed earlier, in general, only small regions of an image contribute its being classified as MA (or PDR in general) or NV (or NPDR in general). Also, the regions that contribute to the classification are in general unavailable, but only the class of the entire image is known (as is the case with all the datasets mentioned in Section III). Therefore, we propose to adopt the MIL framework [8] for the classification of the DR images. In an MIL formulation, one or a subset of instances of features determines the class of the entire bag of features (i.e., the entire image). A classifier built on this approach will divide the feature space into regions where any instance falling on a certain region will decide the class of the bag from that instance.

To support the MIL approach, each image is first cropped to maximally fit a bounding box on the circular retinal region of the fundus image. The image is then quantized into 64 bins by using the quantizer discussed above. The quantized image is then divided into 64 (8X8) equal non-overlapping blocks. Each block is now considered an instance and each image is a bag of instances. AutoCC features for every instance are then extracted using the following algorithm.

*Algorithm for Computing AutoCC:*

*Step 1.* The 64-bin histogram for the instance is recorded

*Step 2.* For any pixel, a neighborhood of 3$X$3 is considered. The pixel value of the center pixel is compared with the pixel value of all the 8 pixels in the neighborhood. A count is made of the number pixels that hold the same value as the pixel under consideration. This count gives us the local spatial distribution of the pixels.

*Step 3.* This count is added to that bin of a spatial distribution histogram of the same 64 bins to which the pixel under consideration belong. This process is repeated for all the pixels in the instance.

*Step 4.* The vector thus formed is divided by the count of global distribution of pixels to get the color auto-correlogram of the instance. This results in a 64 dimensional feature vector for every instance.

Once the features are extracted they are arranged in an instance-bag model and are used to train a multi-class multiple instance learning approach. Now, each block of features can be seen as an instance and each image is a bag. The implementation of the multi-class learning algorithm is based on a multiple instance learning toolbox publicly available at http://www.cs.cmu.edu/~juny/MILL/. We use the "citation KNN" technique proposed by Wang and Zucker [8] to solve the MIL problem. Among all the documented MIL based methods, the authors opt for Citation-KNN because; 1. The feature space is separable by correlation distance, 2. Citation-KNN provides better accuracy than other methods.

### III. EXPERIMENTAL RESULTS

The dataset used to evaluate the features consists of 425 images, assembled manually from well-known databases including DIARETDB0, DIARETDB1 [10], STARE [11] and Messidor (http://messidor.crihan.fr). In total, there are 160 normal images, 181 NPDR images (mostly MAs) and 84 PDR images (mostly NVs).

The proposed features are evaluated against some commonly-used features in the medical imaging literature, such as Gabor features [12] and semantic of neighborhood color moment histogram features (HNM) [13]. These features are used in conjunction with the powerful SVM classifier to form classification schemes. Furthermore, we also consider a state-of-the-art classification scheme in computer vision, SIFT+BoW+SVM, which has been found effective in many computer vision tasks. (Whenever SVM is used, our implementation was based on the well-known LibSVM toolbox [14].) Finally, to provide a case that purely evaluates the benefit of the proposed spectrally-turned AutoCC feature, we also evaluate a scheme using the original AutoCC features in [9] together with the MIL approach in this paper. All these approaches are listed in the first column of Table 1.

To better assess the performance of these various classification schemes on our dataset, we employed 5-fold cross-validation for multiple runs, and then calculated the mean accuracy of classification over all runs. The results are summarized in Table 1. From the table, the proposed approach, which relies on the proposed feature in conjunction with the MIL formulation, achieved 87.6% accuracy, outperforming all the competing approaches by a large margin. This suggests that the proposed method is very promising for DR image classification.

It is interesting to note that, the (generally speaking) powerful approach of "SIFT+BoW+SVM" fares poorly in this study. We made some observations that may explain this. We found that SIFT predominantly captures the optic disk, nerves and some of the edges of the images. It does capture some keypoints but it fails to capture the difference between MA and NV.

To further understand where the misclassified are placed and how the classification is performing with respect to individual classes, the authors have also provided confusion matrices in Table 2 for the "original AutoCC+MIL" scheme and the "spectrally-tuned AutoCC+MIL" scheme respectively. From the matrices, when using the proposed feature with only a 64-dimensional space, the MIL approach performs the classification almost equally well for all the classes. In contrast, the original AutoCC feature has difficulty distinguishing between NPDR and PDR. This illustrates the efficiency of the quantizer designed to spectrally tune the features. Note that, in this specific comparison, the classifier is the same (i.e., MIL) and thus the gain is purely due to the improved feature design.

**Table 1.** Mean accuracy of various methods.

| Approach | Mean Accuracy |
|---|---|
| SIFT+BoW+SVM | 51.14 % |
| Gabor features+SVM | 64.71 % |
| HNM + SVM | 75.76 % |
| Original AutoCC+MIL | 78.01 % |
| Proposed Algorithm | 87.61 % |

**Table 2a.** Confusion matrix for original AutoCC+MIL.

|  | Normal | NPDR | PDR |
|---|---|---|---|
| **Normal** | 79.34 | 14.11 | 6.54 |
| **NPDR** | 5.31 | 92.19 | 2.5 |
| **PDR** | 2.46 | 52.61 | 44.93 |

**Table 2b.** Confusion matrix for the proposed approach.

|  | Normal | NPDR | PDR |
|---|---|---|---|
| **Normal** | 88.12 | 8.11 | 3.77 |
| **NPDR** | 3.05 | 86.05 | 10.9 |
| **PDR** | 1.41 | 8.59 | 90 |

IV. CONCLUSIONS AND FUTURE WORK

In this paper, we proposed a spectrally-tuned AutoCC feature and an MIL framework based on this feature for DR image classification. Experiments with comparison with commonly-used image classification approaches demonstrated that the proposed method is able to achieve significant performance gain. While being potentially promising, the proposed method needs to be further evaluated and improved with larger dataset. This is currently being pursued.

ACKNOWLEDGEMENT: The authors acknowledge support from AHRQ under Grant R21 HS19792-01A1. The views expressed in this paper are purely those of the authors and do not represent official endorsement by AHRQ.